\documentclass[letterpaper]{article} 
\usepackage{aaai25}  
\usepackage{times}  
\usepackage{helvet}  
\usepackage{courier}  
\usepackage[hyphens]{url}  
\usepackage{graphicx} 
\usepackage{natbib}  
\usepackage{caption} 
\newcommand{\ourmethod}{{SEGO}\xspace}

\usepackage{multirow}
\usepackage[ruled,linesnumbered]{algorithm2e}
\usepackage{algpseudocode}
\usepackage{subfigure}
\usepackage{xcolor}
\usepackage{color}
\usepackage[toc,title]{appendix}
\usepackage{enumitem}
\usepackage{float}
\usepackage{amsmath}
\usepackage{amsthm}
\usepackage{booktabs}
\usepackage{amssymb}
\usepackage{adjustbox}
\usepackage{makecell}

\newtheorem{myDef}{Definition}
\newtheorem{thm}{\bf Theorem}
\newtheorem{lemma}{Lemma}

\usepackage{newfloat}
\usepackage{listings}
\DeclareCaptionStyle{ruled}{labelfont=normalfont,labelsep=colon,strut=off} 
\floatstyle{ruled}
\newfloat{listing}{tb}{lst}{}
\floatname{listing}{Listing}
\title{Structural Entropy Guided Unsupervised Graph Out-Of-Distribution Detection}

\author{
    {Yue Hou\textsuperscript{\rm 1,\rm 2}, He Zhu\textsuperscript{\rm 1}, Ruomei Liu\textsuperscript{\rm 1}, Yingke Su\textsuperscript{\rm 2}, Jinxiang Xia\textsuperscript{\rm 1}, Junran Wu\textsuperscript{\rm 1}\thanks{Corresponding authors}, Ke Xu\textsuperscript{\rm 1}}
}
\affiliations{
    \textsuperscript{\rm 1}State Key Laboratory of Complex \& Critical Software Environment, Beihang University\\
    \textsuperscript{\rm 2}Shen Yuan Honors College, Beihang University\\
    \{hou\_yue, roy\_zh, rmliu, suyingke, jinxiangxia, wu\_junran, kexu\}@buaa.edu.cn
}

\begin{document}
\maketitle

\begin{abstract}
With the emerging of huge amount of unlabeled data, unsupervised out-of-distribution (OOD) detection is vital for ensuring the reliability of graph neural networks (GNNs) by identifying OOD samples from in-distribution (ID) ones during testing, where encountering novel or unknown data is inevitable. Existing methods often suffer from compromised performance due to redundant information in graph structures, which impairs their ability to effectively differentiate between ID and OOD data. To address this challenge, we propose SEGO, an unsupervised framework that integrates structural entropy into OOD detection regarding graph classification. Specifically, within the architecture of contrastive learning, SEGO introduces an anchor view in the form of coding tree by minimizing structural entropy. The obtained coding tree effectively removes redundant information from graphs while preserving essential structural information, enabling the capture of distinct graph patterns between ID and OOD samples. Furthermore, we present a multi-grained contrastive learning scheme at local, global, and tree levels using triplet views, where coding trees with essential information serve as the anchor view. Extensive experiments on real-world datasets validate the effectiveness of SEGO, demonstrating superior performance over state-of-the-art baselines in OOD detection. Specifically, our method achieves the best performance on 9 out of 10 dataset pairs, with an average improvement of 3.7\% on OOD detection datasets, significantly surpassing the best competitor by 10.8\% on the FreeSolv/ToxCast dataset pair.
\end{abstract}

\begin{links}
\link{Code}{https://github.com/name-is-what/SEGO}
\end{links}


\section{Introduction}

Out-of-distribution (OOD) detection~\cite{yangbounded,wuenergy,baograph} is a crucial task in machine learning that aims to identify whether a given data point deviates significantly from the training distribution, especially for models deployed in real-world applications where encountering novel or unknown data is inevitable.
In graph-based data, the challenge of OOD detection is heightened due to the complex structure and relationships inherent in graphs. 
In this context, a specific OOD detection model is trained on in-distribution (ID) graphs and then predicts a score for each test sample to indicate its ID/OOD status.

Recent advancements~\cite{wang2024goodat,guo2023data,liu2023good,yuan2024environment} in graph OOD detection and generation have been explored with growing interest.
Several studies~\cite{wang2024goodat,guo2023data} employ well-trained graph neural networks (GNNs)~\cite{gcn_kipf2017semi,gin_xu2019how} to fine-tune OOD detectors to identify OOD samples. However, these methods require annotated ID data to pre-train GNNs, which limits their applicability in scenarios where labeled data is unavailable. 
In contrast, other research~\cite{liu2023good} focuses on training OOD-specific GNN models using only ID data, without relying on any labels or OOD data. They employ unsupervised learning techniques such as graph contrastive learning (GCL) to learn discriminative patterns of unlabeled ID data.


\begin{figure}[t]
    \centering
    \subfigure[ID and OOD graph samples]{
    \includegraphics[width=.95\linewidth]{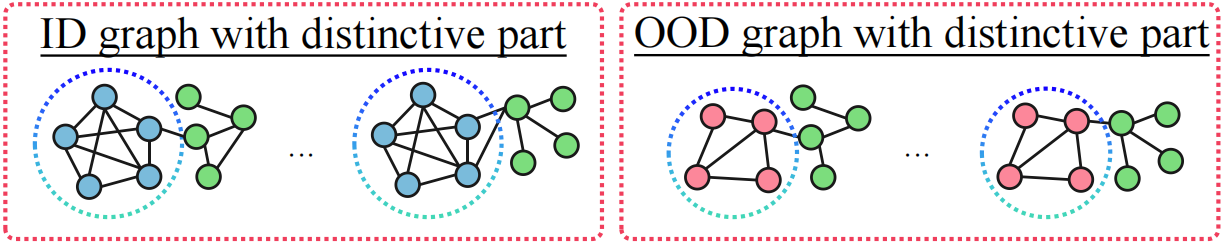}
    }\\
    \vspace{-0.2cm}
        \subfigure[Before minimization]{
        \includegraphics[width=.47\linewidth]{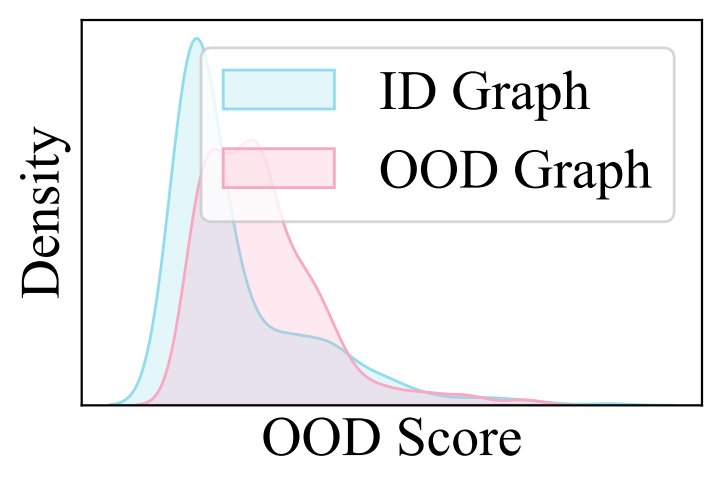}
        }
        \subfigure[After minimization]{
        \includegraphics[width=.47\linewidth]{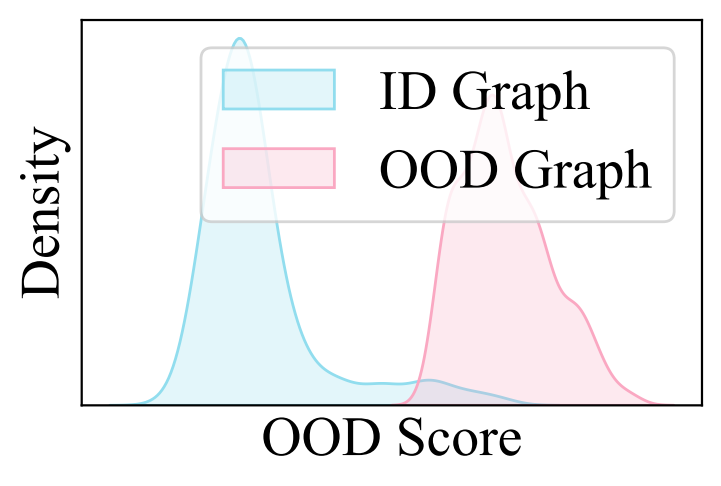}
        }%
\caption{A toy example of ID and OOD graphs and scoring distributions before/after structural entropy minimization.}
\vspace{-1.5em}
\label{fig:toy}
\end{figure}


Despite the progress made in this area, a challenge still remains less explored. Due to the prevalent presence of redundant information in graph structures, current methods struggle to effectively capture and distinguish the essential structure between ID and OOD data.
Without mechanisms to extract substantive information, models are susceptible to irrelevant features and structures that can mislead the learning process. Besides, GCL methods commonly adopt arbitrary augmentations, which may unexpectedly perturb both structural and semantic patterns of the graph, introducing undesired OOD samples and converting ID samples into OOD samples. 
Although methods like GOOD-D~\cite{liu2023good} attempt to mitigate the issue of structural perturbations through perturbation-free data augmentations, they fail to eliminate the interference of irrelevant information.

Structural entropy~\cite{Li2016StructuralEntropy} provides a hierarchical abstraction of graphs to measure the complexity of structure. 
By minimizing structural entropy, the structural uncertainty of the graph is reduced, which aids in capturing essential information and identifying distinct patterns between ID and OOD samples.
We argue that the key to OOD detection is eliminating redundant information in graph structure to focus on the most distinctive and effective information.
Fig.~\ref{fig:toy}(a) presents a toy example of ID and OOD graph data, where the light blue and pink shaded areas represent the distinctive parts (i.e., essential information) in ID and OOD graph structure, respectively. Capturing these distinctive parts of the graph can better differentiate OOD samples from ID graphs.
We also compute the scoring distributions before and after structural entropy minimization on the BZR/COX2 dataset pair (with BZR as ID dataset and COX2 as OOD dataset). As illustrated in Fig.~\ref{fig:toy}(b) and (c), after the minimization, the OOD scores exhibit smaller variance and a decrease in the overlap of scores between OOD and ID samples. The score frequency density plots show that structural entropy minimization effectively removes redundant information in graph samples, preserving the more distinctive parts of the graph, thus enabling the model to detect distributions more effectively.

In this paper, we propose a novel framework, Structural Entropy guided Graph contrastive learning for unsupervised OOD detection, termed \textbf{\ourmethod}, to address this challenge. 
Our approach introduces structural information theory to graph OOD detection for the first time, which can provide significant insights for future research in this field. 
By minimizing structural entropy, our method effectively removes redundant information of the graph while capturing essential structure. 
This allows the model to focus on substantive information that distinguishes ID data from OOD data, improving detection performance.
Specifically, we extract a coding tree from the original graph using structural entropy minimization to obtain redundancy-eliminated structural information. 
Additionally, we theoretically demonstrate that minimizing our contrastive loss preserves the maximum mutual information associated with the ground-truth labels. 
Based on this foundation, we propose a multi-grained contrastive learning scheme using triplet views: the basic view of the original graph, the coding tree as the anchor view representing essential information, and a topological view enriched with topological features. 
Maximum agreement is achieved at local, global, and tree levels, encouraging the model to encode shared information between these views.
Extensive experiments on both real-world datasets demonstrate the superiority of \ourmethod against state-of-the-art (SOTA) baselines. Our method shows an average improvement of 3.7\% in OOD detection across 10 datasets, highlighting its effectiveness in capturing the essential information of graph data for OOD detection tasks.
The main contributions of this work are as follows: 
\begin{itemize}
  \item Guided by structural entropy theory, we propose a novel framework for unsupervised graph OOD detection, termed \ourmethod, which can remove redundant information and capture the essential structure of graphs, significantly improving the model performance. 
  \item To mitigate the information gap between node and graph embeddings, we employ a multi-grained contrastive learning scheme using triplet views, which includes coding tree as an anchor view and operates at local, global, and tree levels.
  \item Extensive experiments validate the effectiveness of \ourmethod, demonstrating superior performance over SOTA baselines in OOD detection.
\end{itemize}

\section{Related Work}
\noindent\textbf{Graph Out-of-distribution Detection. } 
OOD detection aims to identify OOD samples from ID ones and has gained increasing traction due to its wide application for vision~\cite{ssd_sehwag2020ssd,ngc_wu2021ngc,supood1_liang2018enhancing} and language~\cite{nlpood_zhou2021contrastive} data.
OOD detection on graph data can be broadly divided into two categories: graph-level~\cite{liu2023good} and node-level detection~\cite{yangbounded,wuenergy,baograph}.
Lots of existing methods~\cite{zhu2024mario,li2024disentangled,ligraph} focus on improving the generalization ability of GNNs for specific downstream tasks like node classification through supervised learning, rather than identifying OOD samples.
Compared to the works~\cite{supood1_liang2018enhancing,supood2_hendrycks2016baseline} relying on ground-truth labels, relatively less effort has been devoted to unsupervised graph-level OOD detection, which remains an urgent research problem.
In this work, we provide a novel perspective for identifying OOD graphs by focusing on distinctive essential information based on structural entropy.

\noindent\textbf{Graph Contrastive Learning. } 
As an effective graph self-supervised learning paradigm~\cite{ssl_survey_liu2021self,gssl_survey_liu2022graph}, GCL has achieved great success on unsupervised graph representation learning~\cite{dgi_velickovic2019deep,infograph_sun2020infograph,mvgrl_hassani2020contrastive,graphcl_you2020graph,gca_zhu2021graph,gcc_qiu2020gcc,ggd_zheng2022rethinking,ugcl_zheng2022unifying,ding2022structural}. 
Typically, GCL methods involve generating diverse graph views through data augmentation techniques and optimizing the mutual agreement between these views to enhance the representation of samples with similar semantic semantics~\cite{graphcl_you2020graph,gca_zhu2021graph,mvgrl_hassani2020contrastive,ugcl_zheng2022unifying,ding2022structural}.
However, methods that perform augmentation on graph structures may inadvertently introduce undesired OOD samples within ID data, and views that enhance graph features often suffer from containing redundant information. To effectively capture the essential structure of original graphs, this paper introduces a GCL framework guided by structural entropy, innovatively incorporating triplet views and multi-grained contrast.

\noindent\textbf{Structural Entropy. }
Structural entropy~\cite{Li2016StructuralEntropy}, an extension of Shannon entropy~\cite{Shannon}, quantifies system uncertainty by measuring the complexity of graph structures through the coding tree. 
Structural entropy has been widely applied in various domains~\cite{ wu2022structural, wu2022simple,wu2023sega, zhu2023hitin, zhu2024hill,hou2024nc2d}.
In our work, we apply structural entropy in a self-supervised manner to capture the most distinctive part of graphs with essential information for unsupervised graph-level OOD detection.

\section{Notations and Preliminaries}
Before formulating the research problem, we first provide some necessary notations. Let $G=(\mathcal{V},\mathcal{E},\mathbf{X})$ represent a graph, where $\mathcal{V}$ is the set of nodes and $\mathcal{E}$ is the set of edges. The node features are represented by the feature matrix $\mathbf{X} \in \mathbb{R}^{n \times d_f}$, where $n=|\mathcal{V}|$ is the number of nodes and $d_f$ is the feature dimension. The structure information can also be described by an adjacency matrix $\mathbf{A} \in \mathbb{R}^{n \times n}$, so a graph can be alternatively represented by $G=(\mathbf{A},\mathbf{X})$.

\vspace{-1.5mm}
\paragraph{Unsupervised Graph-level OOD Detection. }
We consider an unlabeled ID dataset $\mathcal{D}^{id}=\{G_1^{id}, \cdots, G_{N_1}^{id}\}$ where graphs are sampled from distribution $\mathbb{P}^{id}$ and an OOD dataset $\mathcal{D}^{ood}=\{G_1^{ood}, \cdots, G_{N_2}^{ood}\}$ where graphs are sampled from a different distribution $\mathbb{P}^{ood}$. 
Given a graph $G$ from $\mathcal{D}^{id}$ or $\mathcal{D}^{ood}$, our objective is to detect whether $G$ originates from $\mathbb{P}^{id}$ or $\mathbb{P}^{ood}$. Specifically, we aim to learn a model $f(\cdot)$ that assigns an OOD detection score $s = f(G)$ for an input graph $G$, with a higher $s$ indicating a greater probability that $G$ is from $\mathbb{P}^{ood}$.
The model $f$ is trained solely on the ID dataset $\mathcal{D}^{id}_{train} \subset \mathcal{D}^{id}$ and evaluated on a test set $\mathcal{D}^{id}_{test} \cup \mathcal{D}^{ood}_{test}$ (note that $\mathcal{D}^{id}_{test} \cap \mathcal{D}^{id}_{train} = \emptyset$, $\mathcal{D}^{id}_{test} \subset \mathcal{D}^{id}$, and $\mathcal{D}^{ood}_{test} \subset \mathcal{D}^{ood}$).

\vspace{-1.5mm}
\paragraph{Structural Entropy.} 
Structural entropy is initially proposed \cite{Li2016StructuralEntropy} to measure the uncertainty of graph structure, revealing the essential structure of a graph.
The structural entropy of a given graph ${G}=\left\{\mathcal{V},\mathcal{E}, \mathbf{X}\right\}$ on its coding tree $T$ is defined as:

\vspace{-2.0mm}
\begin{equation}
    \mathcal{H}^T({G})=-\sum\limits_{v_\tau\in T}\frac{g_{v_\tau}}{vol(\mathcal{V})}\log\frac{vol(v_\tau)}{vol(v_\tau^+)},
\end{equation}
where $v_\tau$ is a node in $T$ except for root node and also stands for a subset $\mathcal{V}_\tau \in \mathcal{V}$, $g_{v_\tau}$ is the number of edges connecting nodes in and outside $\mathcal{V}_\tau$, $v_\tau^+$ is the immediate predecessor of of $v_\tau$ and $vol(v_\tau)$, $vol(v_\tau^+)$ and $vol(\mathcal{V})$ are the sum of degrees of nodes in $v_\tau$, $v_\tau^+$ and $\mathcal{V}$, respectively.

\vspace{-1.5mm}
\paragraph{Graph Contrastive Learning.}
In the general graph contrastive learning paradigm for graph classification, two augmented graphs are generated using different graph augmentation operators. Subsequently, representations are generated using a GNN encoder, and further mapped into an embedding space by a shared projection head for contrastive learning. 
A typical graph contrastive loss, InfoNCE~\cite{simclr_chen2020simple,gca_zhu2021graph}, treats the same graph $G_i$ in different views $G_i^\alpha$ and $G_i^\beta$ as positive pairs and other nodes as negative pairs. The graph contrastive learning loss $\mathcal{L}_i $ of graph $G_i$ and total loss $\mathcal{L}$ can be formulated as: 

\vspace{-5.0mm}
\begin{equation}
\vspace{-1.0mm}
\label{eq:contra}
\begin{gathered}
\ell(\mathbf{z}_i^\alpha,\mathbf{z}_i^\beta)  = -\operatorname{log} \frac{e^{sim(\mathbf{z}_i^\alpha,\mathbf{z}_i^\beta)/\tau }}{\sum_{j=1, {j \neq i}}^{N} e^{sim(\mathbf{z}_i^\alpha,\mathbf{z}_j^\alpha)/\tau}+ e^{sim(\mathbf{z}_i^\alpha,\mathbf{z}_j^\beta)/\tau}} , 
\end{gathered}
\vspace{-0.0mm}
\end{equation}

\vspace{-5.0mm}
\begin{equation}
\vspace{-1.0mm}
\label{eq:contra2}
\mathcal{L} = \frac{1}{2N} \sum_{i = 1}^{N} \Big[ \ell(\mathbf{z}_i^\alpha, \mathbf{z}_i^\beta) + \ell(\mathbf{z}_i^\beta, \mathbf{z}_i^\alpha) \Big], 
\vspace{-0.0mm}
\end{equation}

\noindent where $N$ denotes the batch size, $\tau$ is the temperature coefficient, and $sim(\cdot, \cdot)$ stands for cosine similarity function.

\begin{figure*}[ht]
    \includegraphics[width=1\linewidth]{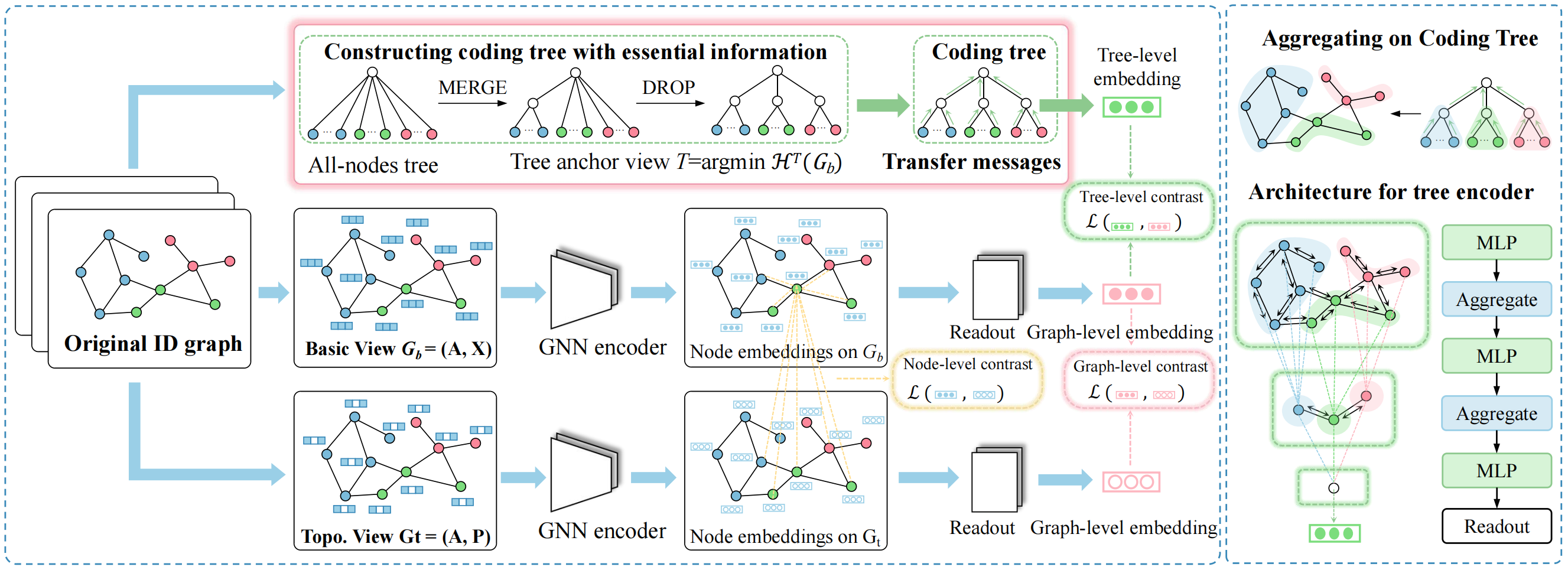}
    \vspace{-2.5mm}
    \caption{Overview of our proposed \ourmethod, which employs multi-grained contrast at local, global, and tree levels using triplet views. 
    The coding tree, obtained by minimizing structural entropy of graph, serves as the essential anchor view that eliminates redundant information. 
    As shown in subfigure on the right, message passing and aggregation on graph are under the guidance of coding tree.
    }
\label{fig:framework}
\vspace{-1em}
\end{figure*}

\section{Methodology}

In this section, we introduce the framework (see Fig.~\ref{fig:framework}), termed \textbf{\ourmethod}. 
We first theoretically reveal that the coding tree with minimum structural entropy effectively captures the essential information of the graph. Additionally, we demonstrate that minimizing our contrastive loss preserves the maximum mutual information associated with ground-truth labels. Based on these insights, we propose a multi-grained contrastive learning using triplet views.

\subsection{Essential View with Redundancy-eliminated Information}
\subsubsection{Redundancy-eliminated Essential Information.}
The key to effectively distinguishing between ID and OOD graphs lies in maximizing the elimination of redundancy while preserving essential information.
According to graph information bottleneck theory (GIB)~\cite{wu2020graph}, retaining important information in a graph view should involve maximizing mutual information (MI) between the output and labels (i.e., $\max I(f(G);y)$) while reducing mutual information between input and output (i.e., $\min I(G; f(G))$). 
For unsupervised downstream tasks where ground-truth labels are unavailable, the objective of minimizing $I(G; f(G))$ is to generate an essential view that retains sufficient information while reducing uncertainty (i.e., redundant information) as much as possible.
This can be expressed as follows:
\begin{equation}
\label{equ: gib}
\resizebox{0.43\textwidth}{!}{$
    \text{GIB: } \max I(f(G); y) - \beta I(G; f(G)) \Rightarrow \min I(G; f(G)),
$}
\end{equation}
where $I(\cdot;\cdot)$ denotes the MI between inputs.

\begin{myDef}
The anchor view with redundancy-eliminated essential information is supposed to be a distinctive substructure of the given graph.
\label{def:view_property}
\end{myDef}

\noindent Now, let $G^\ast$ be the target anchor view  of graph $G$, the mutual information between $G$ and $G^\ast$ can be formulated as:
\begin{equation}
I(G^\ast; G) = \mathcal{H}(G^\ast) - \mathcal{H}(G^\ast|G),
\end{equation}
where $\mathcal{H}(G^\ast)$ is the structural entropy of $G^\ast$ and $\mathcal{H}(G^\ast|G)$ is the conditional entropy of $G^\ast$ conditioned on $G$.

\begin{thm}
The information in $G^\ast$ is a subset of information in $G$ (i.e., $\mathcal{H}(G^\ast) \subseteq \mathcal{H}(G)$); thus, we have:
\begin{equation}
\vspace{-3.0mm}
\mathcal{H}(G^\ast|G)=0.
\end{equation}
\label{theo:hgeq0}
\end{thm}

\noindent Here, the mutual information between $G$ and $G^\ast$ can be rewritten as:
\begin{equation}
I(G^\ast; G) = \mathcal{H}(G^\ast).
\end{equation}
Accordingly, to acquire the anchor view with essential information, we need to optimize:
\begin{equation}
\min\,I(G; f(G)) \Rightarrow \min\,\mathcal{H}(G^\ast).
\label{eq:minH}
\end{equation}
Thus, we argue that the view obtained by minimizing structural entropy of a given graph represents the redundancy-eliminated information, serving as an anchor view that retains the graph's distinctive substructure.

\subsubsection{Maximum Effective Mutual Information.}

With the essential view eliminating redundant information, we also theoretically prove that our \ourmethod effectively captures the maximum mutual information between the representations obtained from the anchor view and labels by minimizing the contrastive loss in Eq.\ref{eq:contra2}.
The InfoMax principle has been widely applied in representation learning literature \cite{Bachman:2019wp,Poole:2019vk,Tschannen:2020uo,yuan2024dynamic}. MI quantifies the amount of information obtained about one random variable by observing the other random variable.
We first introduce two lemmas below.
\begin{lemma}
\label{lm1}
Given that $f$ is a GNN encoder with learnable parameters and $G^\ast$ is the target anchor view of graph $G$. 
If $I(f(G^\ast); f(G))$ reaches its maximum, then $I(f(G^\ast); G)$ will also reach its maximum. 
\end{lemma}

\begin{lemma}
\label{lm2}
Given the anchor view $G^\ast$ of graph $G$ and an encoder $f$, we have
\begin{equation}
I(f(G^\ast); G) \leq I(f(G^\ast); y) + I(G; G^\ast|y).
\end{equation}
\end{lemma}

\begin{thm}
\label{thm:m1}
Optimizing the loss function in Eq.\ref{eq:contra2} is equivalent to maximizing $I(f(G^\ast);f(G))$, leading to the maximization of $I(f(G^\ast); y)$.
\end{thm}
\noindent Theorem \ref{thm:m1} reveals that our anchor view empowers the encoder to acquire enhanced representations through contrastive learning and preserve more information associated with the ground-truth labels, which will boost the performance of downstream tasks.

\vspace{-1.5mm}
\subsection{Redundancy-aware Multi-grained Triplet Contrastive Learning} \label{subsec:hcl}

\subsubsection{Instantiation of Triplet Views.}

Our core idea is to capture the essential information of the training ID data and reduce the interference from irrelevant and redundant information. This allows us to distinguish OOD samples during inference by leveraging the different essential information between ID and OOD data.

We first treat a given graph $G_b = (\mathbf{A}, \mathbf{X})$ as a basic view to directly learn from the original input of the ID data. From this basic view, we construct an anchor view of the graph with minimal structural entropy, as defined by Eq.\ref{eq:minH}. 
According to structural information theory~\cite{Li2016StructuralEntropy}, the structural entropy of a graph needs to be calculated with the coding tree. 
Besides the optimal coding tree with minimum structural entropy (i.e., $\min_{\forall T}\{\mathcal{H}^T(G)\}$), a fixed-height coding tree is often preferred for its better representing the fixed natural hierarchy commonly found in real-world networks. 
Therefore, the $k$-dimensional structural entropy of $G$ is defined using coding tree with fixed height $k$:

\vspace{-1.0mm}
\begin{equation}
    \label{equ: minh}
    \mathcal{H}^{(k)}({G})=\min_{\forall T:\mathrm{Height}(T)=k}\{\mathcal{H}^T({G})\}.
    \vspace{-1.0mm}
\end{equation}

\noindent The total process of generation of a coding tree with fixed height $k$ can be divided into two steps: 1) construction of the full-height binary coding tree and 2) compression of the binary coding tree to height $k$. Given root node $v_r$ of the coding tree $T$, all original nodes in graph ${G}=(\mathcal{V}, \mathcal{E})$ are treated as leaf nodes. 
Correspondingly, based on SEP~\cite{wu2022structural}, we design two efficient operators, $\mathbf{MERGE}$ and $\mathbf{DROP}$, to construct a coding tree $T$ with minimum structural entropy. 
This tree anchor view $T = \arg\min{\mathcal{H}^T({G_b})}$ effectively removes redundant information from graphs while preserving distinctive essential structural information, enabling the capture of distinct graph patterns between ID and OOD samples.

In contrastive learning, traditional graph augmentations such as edge modification or node dropping~\cite{dgi_velickovic2019deep, graphcl_you2020graph} would reduce the model's sensitivity to OOD data. To address this issue, inspired by GOOD-D~\cite{liu2023good}, we adopt a perturbation-free graph augmentation strategy to construct a topological (topo.) view $G_t = (\mathbf{A}, \mathbf{P})$, where $\mathbf{P}$ is a topological matrix formed by concatenating random walk diffusion and Laplacian positional encoding, formally written as $\mathbf{p}_i = [\mathbf{p}_i^{(rw)} \, || \, \mathbf{p}_i^{(lp)}]$.
Specifically, the random walk diffusion encoding is given by $\mathbf{p}_i^{(rw)} = [{RW}_{ii}, {RW}_{ii}^{2}, \cdots, {RW}_{ii}^{r}] \in \mathbb{R}^{r}$, where ${RW} = \mathbf{A}\mathbf{D}^{-1}$ is the random walk transition matrix, and $\mathbf{D}$ is the diagonal degree matrix. The Laplacian positional encoding is defined as $\mathbf{p}_i^{(lp)} = {\Delta}_{ii}$, where ${\Delta} = \mathbf{I} - \mathbf{D}^{-1/2}\mathbf{A}\mathbf{D}^{-1/2}$, with $\mathbf{I}$ being the identity matrix. This strategy ensures discriminative representations for OOD detection.

These triplet views in \ourmethod, namely the basic view $G_b$, tree anchor view $T$, and topo. view $G_t$, collectively integrate multiple levels of information within the graph. Since the coding tree serves as an abstraction of the essential structure on entire graph, maximizing the MI between tree anchor view $T$ and basic view $G_b$ (i.e., $I(T; G_b)$) allows the model to focus on capturing global redundancy-eliminated patterns. 
However, relying solely on $T$ and $G_b$ might result in the overlooking of fine-grained node-level representations, as $I(T; G_b)$ primarily emphasizes coarse-grained graph-level information. Therefore, MI between basic view $G_b$ and topo. view $G_t$ (i.e., $I(G_b; G_t)$) is also required, which is captured in both node and graph embeddings.
The introduction of $G_t$ aligns the information from individual nodes with the overall graph structure, addressing the instability of information. The triplet views in our method mitigate the information gap between node and graph embeddings, effectively capturing both coarse- and fine-grained redundancy-eliminated essential information.

\vspace{-1.5mm}
\subsubsection{Triplet Views Representing Learning.}
To effectively extract embeddings from the basic view $G_b$ and topo view $G_t$, we utilize two parallel and independent GNN encoders (i.e., encoder $f_b$ for the basic view and $f_t$ for the topo view) for representation learning, following the approach in GOOD-D~\cite{liu2023good}.
We employ GIN~\cite{gin_xu2019how} as the backbone for its powerful expression ability. Taking $f_b$ as an example, the propagation in the $l$-th layer can be expressed as,
$\mathbf{h}_{i}^{(b,l)}=\text{MLP}^{(b,l)}\left(\mathbf{h}_{i}^{(b,l-1)}+\sum\nolimits_{v_j \in \mathcal{N}(v_i)} \mathbf{h}_{j}^{(b,l-1)}\right),$
where $\text{MLP}$ is a multilayer perceptron network with 2 layers, $\mathbf{h}_{i}^{(b,l)}$ is the interval embedding of node $v_i$ at the $l$-th layer of encoder $f_b$, $\mathcal{N}(v_i)$ is the set of first-order neighbors of node $v_i$. 
After getting node embeddings, we employ a readout function to acquire the graph embedding:
$\mathbf{h}_{G}^{(b)}=\sum\nolimits_{v_i \in \mathcal{V}_G} \mathbf{h}_{i}^{(b)}$,
where $\mathcal{V}_G$ is the node set of $G$.

For the tree anchor view $T$, the encoder is designed to iteratively transfer messages from the bottom to the top.
Formally, the $l$-th layer of the encoder can be written as, $\mathbf{x}_v^{(l)} = \text{MLP}^{(l)}\left(\sum\nolimits_{u\in\mathcal{C}(v)}\mathbf{x}_u^{(l-1)}\right)$,
where $\mathbf{x}^i_v$ is the feature of $v$ in the $i$-th layer of coding tree $T$, $\mathbf{x}^0_v$ is the input feature of leaf nodes, and $\mathcal{C}(v)$ refers to the children of $v$.
Once the features reach the root node, a readout function is applied to obtain the tree-level embedding $\mathbf{z}_{T}$.

\vspace{-1.5mm}
\subsubsection{Multi-grained Contrastive Learning Objectives.} 

To capture the multi-grained mutual information between views, we employ a multi-grained contrastive learning scheme that extracts features at three distinct levels: the \textbf{local-level} for fine-grained feature extraction, the \textbf{global-level} for coarse-grained feature extraction, and the \textbf{tree-level} for capturing essential information of the entire graph. 
To maximize the agreement between node embeddings from different views of the same graph, we first map $\mathbf{h}_{i}^{(b)}$ and $\mathbf{h}_{i}^{(t)}$ into node-space embeddings $\mathbf{z}_{i}^{(b)}$ and $\mathbf{z}_{i}^{(t)}$ using MLP-based projection networks.
The local-level contrast focuses on both inter-view and intra-view node relationships, defined as follows:
\begin{equation}
\vspace{-1.2mm}
\label{eq:loss_node}
\mathcal{L}_{local} = \frac{1}{|\mathcal{B}|} \sum\limits_{G_j \in \mathcal{B}} \frac{1}{2|\mathcal{V}_{j}|} \sum\limits_{v_i \in \mathcal{V}_{j}} \left[ \ell(\mathbf{z}_{i}^{(b)}, \mathbf{z}_{i}^{(t)}) + \ell(\mathbf{z}_{i}^{(t)}, \mathbf{z}_{i}^{(b)}) \right],
\vspace{-0.0mm}
\end{equation}

\noindent where $\mathcal{B}$ is a training batch containing multiple graph samples, $\mathcal{V}_{j}$ is the node set of graph $G_j$, $\ell(\mathbf{z}_{i}^{(t)},\mathbf{z}_{i}^{(b)})$ and $\ell(\mathbf{z}_{i}^{(b)},\mathbf{z}_{i}^{(t)})$ are calculated following Eq.\ref{eq:contra}.

\noindent The global-level contrast allows the model to identify coarse-grained patterns that might be overlooked when focusing solely on finer details:
\begin{equation}
\vspace{-1.8mm}
\label{eq:loss_graph}
\mathcal{L}_{global} = \frac{1}{2|\mathcal{B}|} \sum_{G_i \in \mathcal{B}} \Big[ \ell(\mathbf{z}_{G_i}^{(b)},\mathbf{z}_{G_i}^{(t)}) + \ell(\mathbf{z}_{G_i}^{(t)},\mathbf{z}_{G_i}^{(b)}) \Big],
\vspace{-0.0mm}
\end{equation}
\noindent where $\mathbf{z}_{G}^{(b)}$ and $\mathbf{z}_{G}^{(t)}$ are transformed from $\mathbf{h}_{G}^{(b)}$ and $\mathbf{h}_{G}^{(t)}$ using MLP-based projection networks. 

\noindent Tree-level contrast operates at a higher level of abstraction, which can be calculated by: 
\begin{equation}
\vspace{-1.5mm}
\label{eq:loss_tree}
\mathcal{L}_{tree} = \frac{1}{2|\mathcal{B}|} \sum_{G_i \in \mathcal{B}} \Big[ \ell(\mathbf{z}_{G_i}^{(b)},\mathbf{z}_{T_i}) + \ell(\mathbf{z}_{T_i},\mathbf{z}_{G_i}^{(b)}) \Big].
\vspace{-0.0mm}
\end{equation}

\begin{table*}[t]
\resizebox{1\textwidth}{!}{
\begin{tabular}{l | cccccccccc|cc}
\toprule
ID dataset & BZR & PTC-MR & AIDS & ENZYMES & IMDB-M & Tox21 & FreeSolv & BBBP & ClinTox & Esol & \multirow{2}{1.8em}{A.A.} & \multirow{2}{1.8em}{A.R.} \\
\cmidrule{1-11}
OOD dataset & COX2 & MUTAG & DHFR & PROTEIN & IMDB-B & SIDER & ToxCast & BACE & LIPO & MUV \\
\midrule
PK-LOF      & 42.22±8.39 & 51.04±6.04 & 50.15±3.29 & 50.47±2.87 & 48.03±2.53 & 51.33±1.81 & 49.16±3.70 & 53.10±2.07 & 50.00±2.17 & 50.82±1.48 & 49.63 & 12.9\\
PK-OCSVM    & 42.55±8.26 & 49.71±6.58 & 50.17±3.30 & 50.46±2.78 & 48.07±2.41 & 51.33±1.81 & 48.82±3.29 & 53.05±2.10 & 50.06±2.19 & 51.00±1.33 & 49.52 & 12.8 \\
PK-iF & 51.46±1.62 & 54.29±4.33 & 51.10±1.43 & 51.67±2.69 & 50.67±2.47 & 49.87±0.82 & 52.28±1.87 & 51.47±1.33 & 50.81±1.10 & 50.85±3.51 & 51.45 & 11.1 \\
WL-LOF   & 48.99±6.20 & 53.31±8.98 & 50.77±2.87 & 52.66±2.47 & 52.28±4.50 & 51.92±1.58 & 51.47±4.23 & 52.80±1.91 & 51.29±3.40 & 51.26±1.31 & 51.68 & 10.4 \\
WL-OCSVM    & 49.16±4.51 & 53.31±7.57 & 50.98±2.71 & 51.77±2.21 & 51.38±2.39 & 51.08±1.46 & 50.38±3.81 & 52.85±2.00 & 50.77±3.69 & 50.97±1.65 & 51.27 & 11.1 \\
WL-iF  & 50.24±2.49 & 51.43±2.02 & 50.10±0.44 & 51.17±2.01 & 51.07±2.25 & 50.25±0.96 & 52.60±2.38 & 50.78±0.75 & 50.41±2.17 & 50.61±1.96 & 50.87 & 12.4 \\
\midrule
OCGIN  & 76.66±4.17 & 80.38±6.84 & 86.01±6.59 & 57.65±2.96 & 67.93±3.86 & 46.09±1.66 & 59.60±4.78 & 61.21±8.12 & 49.13±4.13 & 54.04±5.50 & 63.87 & 7.9 \\
GLocalKD  & 75.75±5.99 & 70.63±3.54 & 93.67±1.24 & 57.18±2.03 & 78.25±4.35 & 66.28±0.98 & 64.82±3.31 & 73.15±1.26 & 55.71±3.81 & 86.83±2.35 & 72.23 & 5.1 \\
\midrule

InfoGraph-iF & 63.17±9.74 & 51.43±5.19 & 93.10±1.35 & 60.00±1.83 & 58.73±1.96 & 56.28±0.81 & 56.92±1.69 & 53.68±2.90 & 48.51±1.87 & 54.16±5.14 & 59.60 & 8.5 \\
InfoGraph-MD & 86.14±6.77 & 50.79±8.49 & 69.02±11.67 & 55.25±3.51 & \textbf{81.38±1.14} & 59.97±2.06 & 58.05±5.46 & 70.49±4.63 & 48.12±5.72 & 77.57±1.69 & 65.68 & 7.4 \\
GraphCL-iF & 60.00±3.81 & 50.86±4.30 & 92.90±1.21 & 61.33±2.27 & 59.67±1.65 & 56.81±0.97 & 55.55±2.71 & 59.41±3.58 & 47.84±0.92 & 62.12±4.01 & 60.65 & 8.7 \\
GraphCL-MD & 83.64±6.00 & 73.03±2.38 & 93.75±2.13 & 52.87±6.11 & 79.09±2.73 & 58.30±1.52 & 60.31±5.24 & 75.72±1.54 & 51.58±3.64 & 78.73±1.40 & 70.70 & 5.3 \\
 
GOOD-D$_{simp}$ & 93.00±3.20 & 78.43±2.67 & 98.91±0.41 & \underline{61.89±2.51} & 79.71±1.19 & 65.30±1.27 & 70.48±2.75 & 81.56±1.97 & 66.13±2.98 & 91.39±0.46 & 78.68 & 3.2 \\
GOOD-D & \underline{94.99±2.25} & \underline{81.21±2.65} & \underline{99.07±0.40} & 61.84±1.94 & 79.94±1.09 & \underline{66.50±1.35} & \underline{80.13±3.43} & \underline{82.91±2.58} & \underline{69.18±3.61} & \underline{91.52±0.70} & \underline{80.73} & \underline{2.2} \\
\midrule 
\ourmethod & \textbf{96.66±0.91} & \textbf{85.02±0.94} & \textbf{99.48±0.11} & \textbf{64.42±4.95} & \underline{80.27±0.92} & \textbf{66.67±0.82} & \textbf{90.95±1.93} & \textbf{87.55±0.13} & \textbf{78.99±2.81} & \textbf{94.59±0.94} & \textbf{84.46} & \textbf{1.1} \\

\bottomrule
\end{tabular}}
\vspace{-0.2cm}
\centering
\caption{OOD detection results in terms of AUC ($\%$, mean $\pm$ std). The best and runner-up results are highlighted with \textbf{bold} and \underline{underline}, respectively. A.A. is short for average AUC. A.R. implies the abbreviation of average rank. The results of baselines are derived from the published works.} 
\label{tab:main_od}
\vspace{-0.4cm}
\end{table*}

During the training phase, we introduce the standard deviation of prediction errors to adaptively adjust the balance of local and global information. This strategy automatically allocates the weights for loss and score terms. 
Concretely, the overall loss is calculated by:

\vspace{-3.0mm}
\begin{equation}
\label{eq:loss_all}
\mathcal{L} = \mathcal{L}_{tree} + \sigma_{l}^{\theta} \mathcal{L}_{local} + \sigma_{g}^{\theta} \mathcal{L}_{global},
\vspace{-0.0mm}
\end{equation}

\noindent where $\sigma_{l}$ and $\sigma_{g}$ are the standard deviations of predicted errors of the node and graph levels, respectively, and $\theta \geq 0$ is a hyper-parameter that controls the strength of self-adaptiveness, penalizing the term with a larger deviation.
During the inference phase, to balance the scores of different levels, we employ z-score normalization based on the mean values and standard deviations of the predicted errors of training samples:
$s_{G_i} = \frac{s_{l} - \mu_{l}}{\sigma_{l}} + \frac{s_{g} - \mu_{g}}{\sigma_{g}}$,
where $\mu_{l}$ and $\mu_{g}$ represent the mean values of the predicted errors at the corresponding levels for the training samples.

\section{Experiment}

In this section, we empirically evaluate the effectiveness of the proposed \ourmethod. 
In particular, the experiments are unfolded by answering the following research questions:

\begin{itemize}[leftmargin=*,noitemsep,topsep=1.5pt]
    \item \textbf{RQ1:} How effective is \ourmethod compared with competitive baselines on identifying OOD graphs?
    \item \textbf{RQ2:} How transferable is \ourmethod to anomaly detection?
    \item \textbf{RQ3:} How do our multi-grained contrastive losses affect SEGO’s performance?
    \item \textbf{RQ4:} How about the parameter sensitivity of \ourmethod?
\end{itemize}

\vspace{-1.5mm}
\subsection{Experimental Setups}
\noindent \textbf{Datasets.}
For OOD detection, we employ 10 pairs of datasets from two mainstream graph data benchmarks (i.e., TUDataset~\cite{tu_Morris2020} and OGB~\cite{ogb_hu2020open}) following GOOD-D~\cite{liu2023good}. 
We also conduct experiments on anomaly detection settings, where 5 datasets from TUDataset~\cite{tu_Morris2020} are used for evaluation, where the samples in minority class or real anomalous class are viewed as anomalies, while the rest are as normal data. 

\noindent \textbf{Baselines.}
We compare \ourmethod with 14 competing baseline methods, including 6 GCL~\cite{infograph_sun2020infograph,graphcl_you2020graph,liu2023good} based methods,
6 graph kernel based methods~\cite{gk_vishwanathan2010graph,wlgk_shervashidze2011weisfeiler}, 
and 2 end-to-end graph anomaly detection methods~\cite{ocgin_zhao2021using,glocalkd_ma2022deep}.

\noindent \textbf{Evaluation and Implementation. }
We evaluate \ourmethod with a popular OOD detection metric, i.e., area under receiver operating characteristic Curve (AUC). Higher AUC values indicate better performance. 
The reported results are the mean performance with standard deviation after 5 runs.

\vspace{-1.5mm}
\subsection{Performance on OOD Detection (RQ1)} \label{subsec:od_results}

To answer RQ1, we compare our proposed methods with 14 competing methods in OOD detection tasks. The AUC results are reported in Table~\ref{tab:main_od}. From the comparison results, we observe that \ourmethod achieves superior performance improvements over the baselines. Specifically, \ourmethod achieves the best performance on 9 out of 10 dataset pairs and ranks first on average among all baselines with an average rank (A.R.) of 1.1.
Additionally, \ourmethod outperforms all the compared methods in terms of average AUC with a score of 84.46, which is 3.7\% higher than the second-best method GOOD-D~\cite{liu2023good}. Notably, on the FreeSolv/ToxCast dataset pair, \ourmethod surpasses the best competitor by 10.8\%.
Although \ourmethod nearly achieves optimal results on the IMDB-M/IMDB-B datasets, it falls short likely because its coding tree only approximates and doesn't fully remove redundant information. Additionally, the high connectivity and edge density of social network datasets introduce more redundancy, making it harder to capture essential information.
These results underscore the superiority of \ourmethod in OOD detection tasks, demonstrating its ability to capture essential information across different granular levels.

\begin{table}[t]
\centering
\vspace{-0.3cm}
\resizebox{\linewidth}{!}{
\begin{tabular}{l | ccccc}
\toprule
Dataset & ENZYMES & DHFR & BZR & NCI1 & IMDB-B \\
\midrule
PK-OCSVM & 53.67±2.66 & 47.91±3.76 & 46.85±5.31 & 49.90±1.18 & 50.75±3.10 \\
PK-iF & 51.30±2.01 & 52.11±3.96 & 55.32±6.18 & 50.58±1.38 & 50.80±3.17 \\
WL-OCSVM & 55.24±2.66 & 50.24±3.13 & 50.56±5.87 & 50.63±1.22 & 54.08±5.19 \\
WL-iF & 51.60±3.81 & 50.29±2.77 & 52.46±3.30 & 50.74±1.70 & 50.20±0.40 \\
GraphCL-iF & 53.60±4.88 & 51.10±2.35 & 60.24±5.37 & 49.88±0.53 & 56.50±4.90 \\
OCGIN & 58.75±5.98 & 49.23±3.05 & 65.91±1.47 & \textbf{71.98±1.21} & 60.19±8.90 \\
GLocalKD & 61.39±8.81 & 56.71±3.57 & 69.42±7.78 & {68.48±2.39} & 52.09±3.41 \\
GOOD-D$_{simp}$ & 61.23±4.58 & \underline{62.71±3.38} & 74.48±4.91 & 59.56±1.62 & 65.49±1.06 \\
GOOD-D & \underline{63.90±3.69} & 62.67±3.11 & \underline{75.16±5.15} & 61.12±2.21 & \underline{65.88±0.75} \\
\midrule
\ourmethod & \textbf{76.62±7.35} & \textbf{65.31±2.98} & \textbf{89.21±4.51} & \underline{70.34±1.31} & \textbf{66.48±0.38} \\
\bottomrule
\end{tabular}}
\caption{Anomaly detection results in terms of AUC ($\%$, mean $\pm$ std). The best and runner-up results are highlighted with \textbf{bold} and \underline{underline}, respectively.} 
\label{tab:ad}
\vspace{-0.3cm}
\end{table}

\begin{table*}[t]
\centering
\resizebox{1\textwidth}{!}{
\begin{tabular}{ccc | cccccccccc}
\toprule
\multirow{2}{*}{$\mathcal{L}_{tree}$} & \multirow{2}{*}{$\mathcal{L}_{global}$} & \multirow{2}{*}{$\mathcal{L}_{local}$} & BZR & PTC-MR & AIDS & ENZYMES & IMDB-M & Tox21 & FreeSolv & BBBP & ClinTox & Esol \\
\cmidrule{4-13}
  &   &   & COX2 & MUTAG & DHFR & PROTEIN & IMDB-B & SIDER & ToxCast & BACE & LIPO & MUV \\
\midrule
\checkmark & -& - & 54.79±4.08 & 58.20±3.87 & 43.68±7.36 & 49.26±1.11 & 49.56±5.76 & 49.26±5.10 & 49.89±2.95 & 50.53±0.63 & 51.97±4.58 & 54.49±3.57 \\
- & \checkmark& -  & \underline{87.44±4.66} & 77.84±3.71 & 97.60±1.05 & 56.74±1.96 & 75.22±1.91 & 65.07±1.32 & 78.40±6.44 & 77.66±2.29 & 70.11±2.44 & 89.57±2.80 \\
- & -& \checkmark  & 83.51±4.14 & 72.48±3.77 & 96.84±0.58 & 60.85±2.95 & \underline{79.34±1.81} & 62.58±0.67 & 59.48±2.20 & 69.53±2.29 & 53.29±4.32 & 86.49±1.20 \\
\checkmark & \checkmark& -  & 87.27±8.21 & \textbf{87.71±1.35} & 97.97±0.04 & 54.82±2.74 & 74.51±1.52 & 64.84±0.29 & \underline{89.34±0.06} & \textbf{88.34±1.64} & \textbf{79.21±4.55} & \underline{94.13±1.32} \\
\checkmark & -& \checkmark  & 79.36±8.69 & 55.08±1.29 & 90.66±3.40 & \underline{63.38±4.18} & 72.96±3.73 & 55.68±2.67 & 61.01±5.29 & 70.13±0.26 & 52.14±2.58 & 77.78±1.01 \\
- & \checkmark& \checkmark  & 86.29±1.09 & 77.53±4.03 & \underline{98.23±0.19} & 61.55±1.47 & 75.27±0.54 & \underline{65.44±1.14} & 88.04±1.15 & 80.43±2.58 & 65.89±4.58 & 90.94±1.17 \\
\midrule
\checkmark & \checkmark& \checkmark 
& \textbf{96.66±0.91} & \underline{85.02±0.94} & \textbf{99.48±0.11} & \textbf{64.42±4.95} & \textbf{80.27±0.92} & \textbf{66.67±0.82} & \textbf{90.95±1.93} & \underline{87.55±0.13} & \underline{78.99±2.81} & \textbf{94.59±0.94} \\
\bottomrule
\end{tabular}}
\vspace{-0.2cm}
\caption{Ablation study results of \ourmethod and its variants in terms of AUC ($\%$, mean $\pm$ std).} 
\label{tab:ablation}
\vspace{-0.2cm}
\end{table*}

\vspace{-1.5mm}
\subsection{Performance on Anomaly Detection (RQ2)}

To investigate if \ourmethod can generalize to the anomaly detection setting~\cite{ocgin_zhao2021using,glocalkd_ma2022deep}, we conduct experiments on 5 datasets following the benchmark in GLocalKD and GOOD-D~\cite{glocalkd_ma2022deep,liu2023good}, where only normal data are used for model training. 
The results are shown in Table~\ref{tab:ad}.
From the results, we find that \ourmethod shows significant performance improvements compared to other baseline methods. 
Capturing common patterns in the anomaly detection setting is crucial, which is directly reflected in the performance. 
Thus, we can conclude that \ourmethod indeed has a strong capability to learn the essential information of normal graph data.

\vspace{-1.5mm}
\subsection{Ablation Study (RQ3)} \label{subsec:abla}
\noindent \textbf{Ablation of Multi-grained Contrastive Loss.}
To address RQ3, we conducted ablation experiments on the OOD detection task by separately removing the different levels of contrastive losses, namely node-, graph-, and tree-level losses. The results are presented in Table~\ref{tab:ablation}.
Firstly, we observe that applying contrastive loss across all three levels (the last row) achieves the best results on 7 out of 10 datasets and shows promising performance on the remaining datasets. 
This further elucidates that \ourmethod better captures the essential information, leading to superior performance in most OOD detection scenarios.
Notably, we notice that removing the local-level contrast $\mathcal{L}_{local}$ improves performance on certain dataset pairs (e.g., PTC-MR/MUTAG, BBBP/BACE, and ClinTox/LIPO). These datasets primarily consist of biomolecular data, where molecular activity and interactions are more dependent on the overall topological structure rather than individual node features. Removing $\mathcal{L}_{local}$, which focuses on node-level feature extraction and optimization, reduces interference and allows the model to focus more on the global graph structure, enhancing performance on these datasets.

\begin{figure}[!t]
\begin{center}
\centerline{\includegraphics[width=\linewidth]{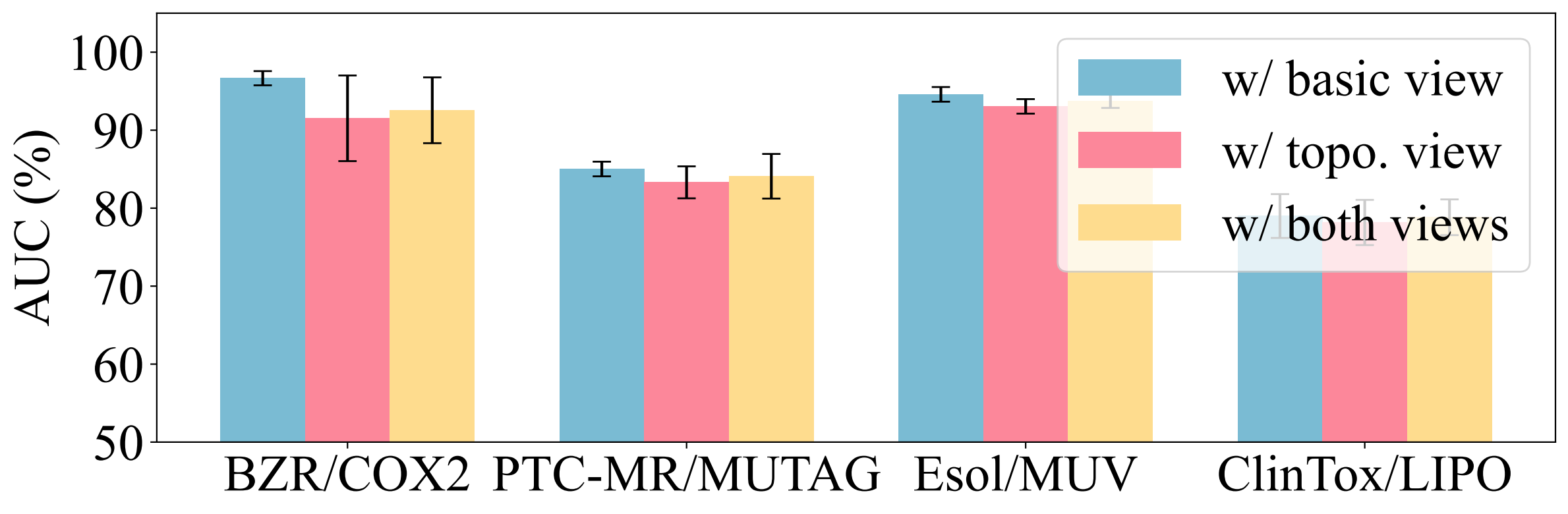}}
\vspace{-0.1cm}
\caption{The effectiveness of different views.}
\label{fig:view}
\end{center}
\vspace{-0.5cm}
\end{figure}

\noindent \textbf{Effectiveness of MI in Triplet Views. }\ourmethod utilizes $I(T; G_b)$ at the tree-level contrastive loss. Here, we explore the effectiveness of MI between the anchor view $T$ and topo. view ($I(T; G_t)$), as well as both views ($I(T; G_b)\cup I(T; G_t)$) in identifying OOD graphs.
As shown in Fig.~\ref{fig:view}, we observe that using $I(T; G_b)$ is more effective in eliminating redundancy from the original graph, whereas, in the view $G_t$, data augmentation reintroduces redundant information, leading to sub-optimal performance.

\begin{figure}[t!]
 \centering
 \vspace{-0.1cm}
 \hspace{-0.18cm}
 \subfigure[w.r.t $\mathbf{z}_{G}^{(b)}$]{
   \includegraphics[width=0.31\linewidth]{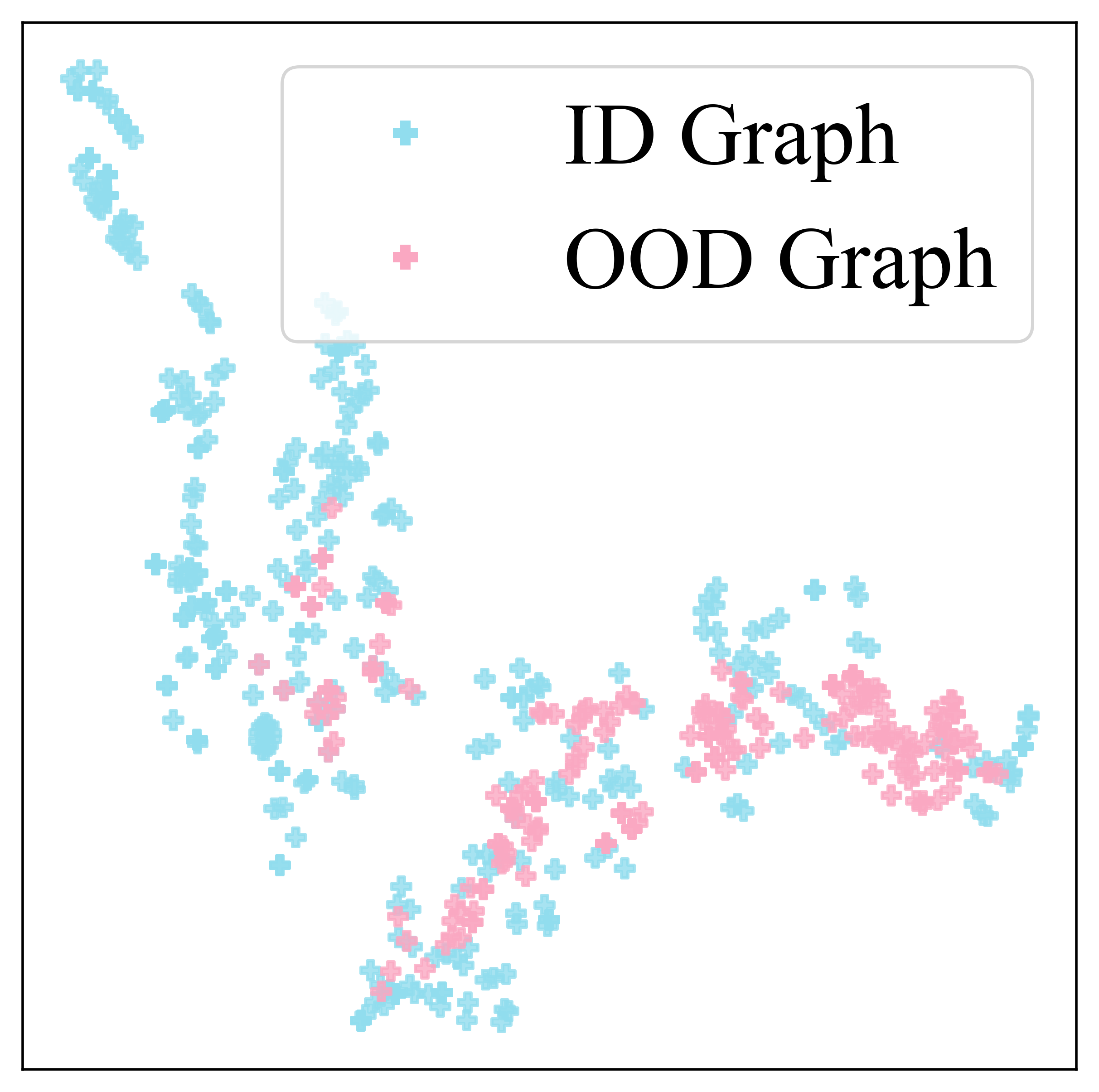}
   \label{subfig:graph_f}}
 \hspace{-0.18cm}
 \subfigure[w.r.t $\mathbf{z}_{G}^{(t)}$]{
   \includegraphics[width=0.31\linewidth]{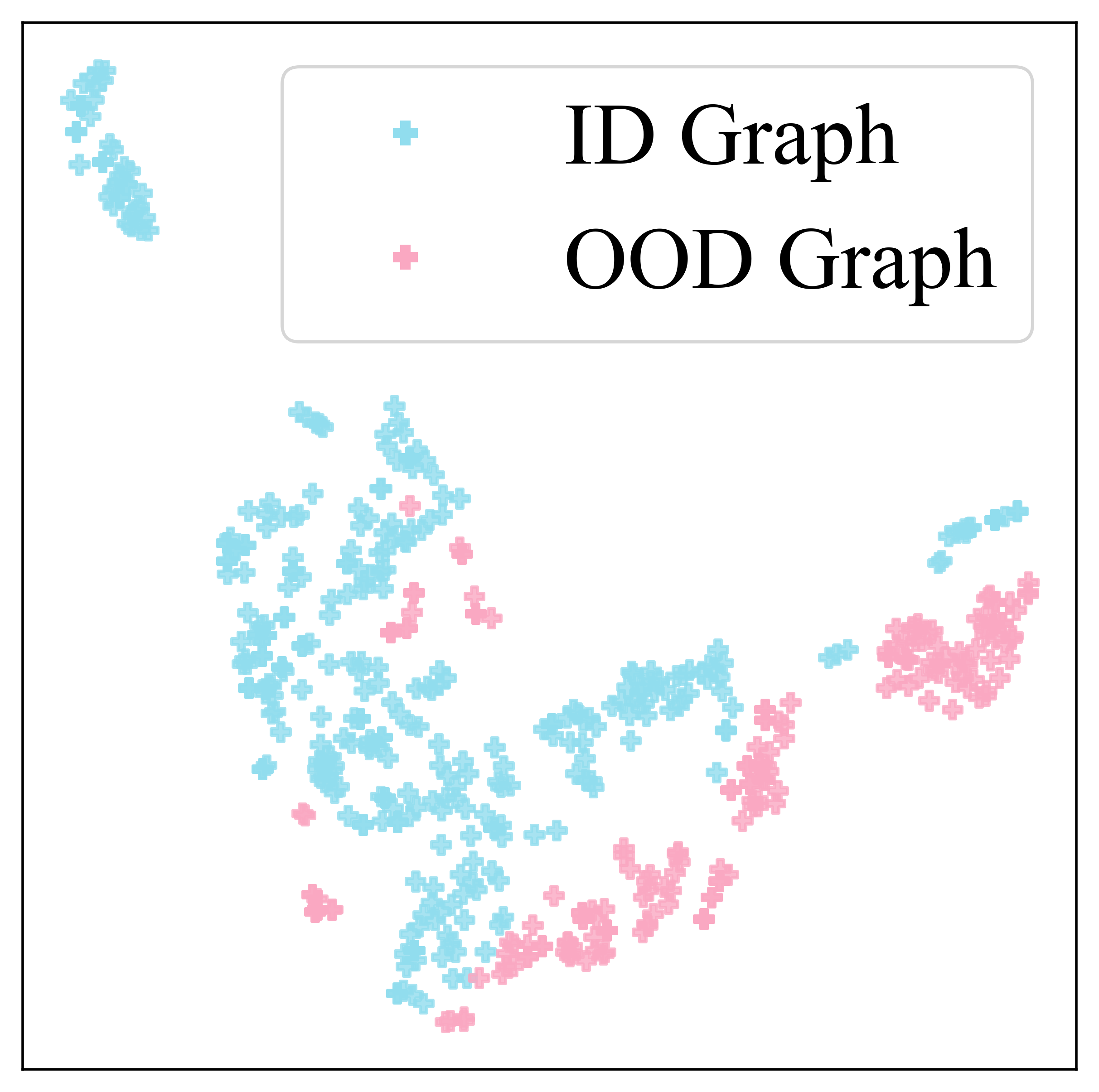}
   \label{subfig:graph_s}} 
 \hspace{-0.18cm}
 \subfigure[w.r.t $\mathbf{z}_{T}$]{
   \includegraphics[width=0.31\linewidth]{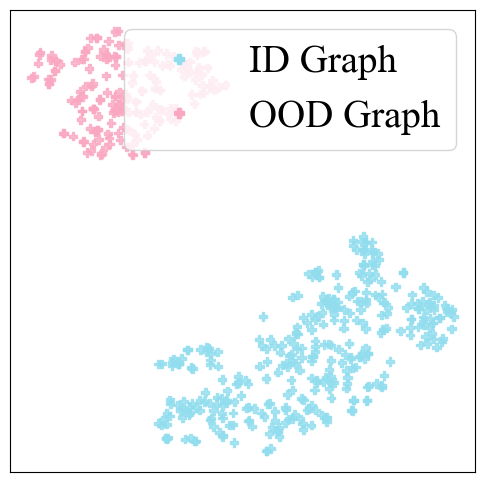}
   } 
 \vspace{-0.2cm}
  \caption{T-SNE visualization of embeddings.}
  \vspace{-0.3cm}
 \label{fig:v}
\end{figure}

\noindent \textbf{Visualization. }
We also visualize the embeddings on PTC-MR/MUTAG dataset pair learned by \ourmethod in triplet views via t-SNE~\cite{tsne_van2008visualizing}.
As shown in Fig.~\ref{fig:v}(a)-(c), the embeddings of ID and OOD graphs are well-separated across these views. Among them, the representation gap in $\mathbf{z}_{T}$ is the most pronounced, highlighting the effectiveness of coding tree in extracting essential structures.

\vspace{-1.5mm}
\subsection{Parameter Study (RQ4)} \label{subsec:param}

\begin{figure}[!t]
\vspace{-0.2cm}
\begin{center}
\centerline{\includegraphics[width=\linewidth]{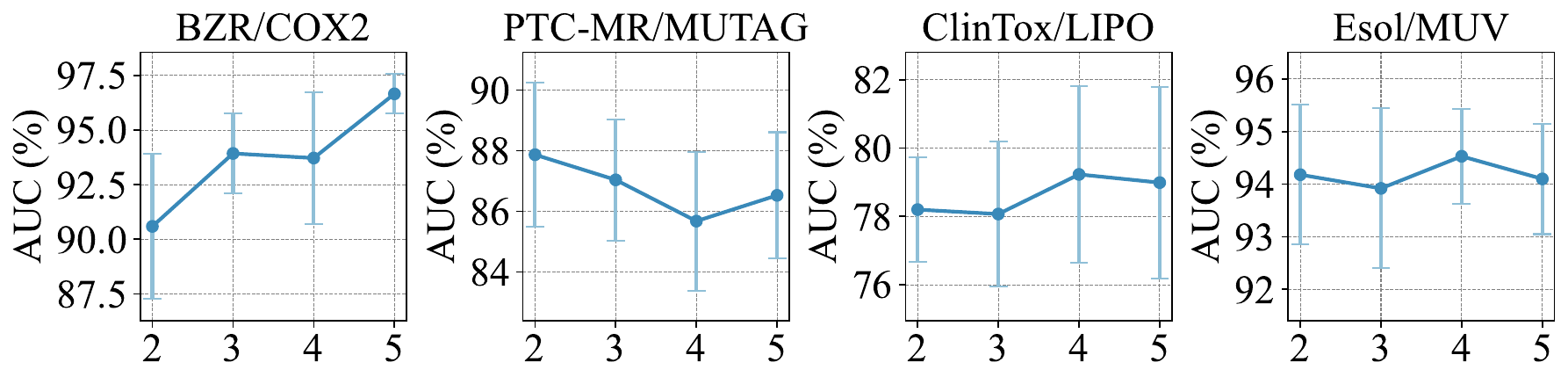}}
\vspace{-0.2cm}
\caption{The natural hierarchy of graph.}
\label{fig:h}
\end{center}
\vspace{-0.5cm}
\end{figure}

\noindent \textbf{The Height $k$ of Graph's Natural Hierarchy. }
In the experimental setup, the height $k$ of the coding tree is consistently set to 5, in alignment with the GNN encoder. Here, we delve deeper into the effect of the height $k$ on the graph's natural hierarchy. The specific performance of \ourmethod under each height $k$, ranging from 2 to 5, on OOD detection is shown in Fig.~\ref{fig:h}. 
We can observe that the optimal height $k$ with the highest accuracy varies among datasets.

\begin{figure}[!t]
\begin{center}
\centerline{\includegraphics[width=1\linewidth]{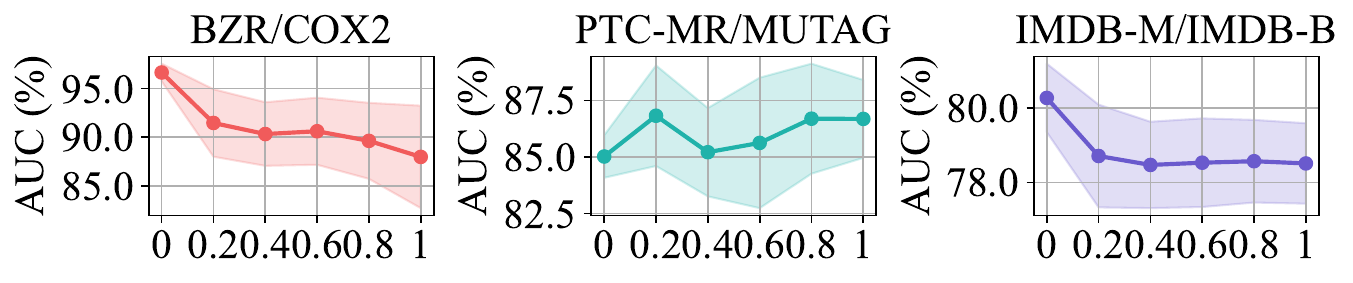}}
\caption{The sensitivity of self-adaptiveness strength $\theta$.}
\label{fig:a}
\end{center}
\vspace{-0.6cm}
\end{figure}

\noindent \textbf{Self-adaptiveness Strength $\theta$. } 
To analyze the sensitivity of $\theta$ for \ourmethod, we alter the value of $\theta$ from $0$ to $1$. The AUC w.r.t different selections of $\theta$ is plotted in Fig.~\ref{fig:a}. 
We can observe that the variation in AUC with changes in $\theta$ is not entirely consistent across different datasets. This aligns with the findings from the ablation study in Section~\ref{subsec:abla}, where we noted that the essential information carried by different datasets varies in their dependence on local node information versus global graph information.

\section{Conclusion}
In this paper, we make the first attempt to introduce structural information theory 
into unsupervised OOD detection regarding graph classification.
For this task, we propose a novel structural entropy guided graph contrastive learning framework, termed \ourmethod, that minimizes structural entropy to capture essential graph information while removing redundant information. 
Our \ourmethod employs a multi-grained contrastive learning at node, graph, and tree levels with triplet views, including a coding tree with minimum structural entropy as the anchor view.
Extensive experiments on real-world datasets validate the effectiveness of \ourmethod, demonstrating superior performance over state-of-the-art baselines.

\section{Acknowledgments}
This work has been supported by the Guangxi Science and Technology Major Project, China (No. AA22067070), NSFC (Grant No. 61932002) and CCSE project (CCSE-2024ZX-09).
\bibliography{main}

\end{document}